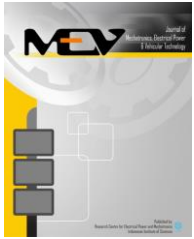

# Journal of Mechatronics, Electrical Power, and Vehicular Technology

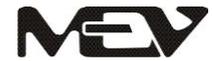

# Experimental review of distance sensors for indoor mapping

Midriem Mirdanies [a, *], Roni Permana Saputra [a, b]

[a] *Research Centre for Electrical Power and Mechatronics, Indonesian Institute of Sciences*
*Komp. LIPI Bandung, Jl. Sangkuriang, Gd. 20, Lt. 2, Bandung 40135, Indonesia*
[b] *Dyson School of Design Engineering, Imperial College London,*
*10 Princes Gardens, South Kensington, London, United Kingdom*

**Abstract**

One of the most important required ability of a mobile robot is perception. An autonomous mobile robot has to be able to gather information from the environment and use it for supporting the accomplishing task. One kind of sensor that essential for this process is distance sensor. This sensor can be used for obtaining the distance of any objects surrounding the robot and utilize the information for localizing, mapping, avoiding obstacles or collisions and many others. In this paper, some of the distance sensor, including Kinect, Hokuyo UTM-30LX, and RPLidar were observed experimentally. Strengths and weaknesses of each sensor were reviewed so that it can be used as a reference for selecting a suitable sensor for any particular application. A software application has been developed in C programming language as a platform for gathering information for all tested sensors. According to the experiment results, it showed that Hokuyo UTM-30LX results in random normally distributed error on measuring distance with average error 21.94 mm and variance 32.11. On the other hand, error measurement resulted by Kinect and RPLidar strongly depended on measured distance of the object from the sensors, while measurement error resulted by Kinect had a negative correlation with the measured distance and the error resulted by RPLidar sensor had a positive correlation with the measured distance. The performance of these three sensors for detecting a transparent object shows that the Kinect sensors can detect the transparent object on its effective range measurement, Hokuyo UTM-30LX can detect the transparent object in the distance more than equal to 200 mm, and the RPLidar sensor cannot detect the transparent object at all tested distance. Lastly, the experiment shows that the Hokuyo UTM-30LX has the fastest processing time significantly, and the RPLidar has the slowest processing time significantly, while the processing time of Kinect sensor was in between. These processing times were not significantly affected by various tested distance measurement.

## I. Introduction

Research and development on a mobile robot that has an ability to accomplish the required task without human intervention (i.e., autonomous system) have attracted many researchers in robotics and mechatronics research field in the recent years. To operate it autonomously, it is essential for a mobile robot to have an ability to percept itself and the surrounding environment. One of the important sensors for this operation is distance sensor. On mobile robot application, distance sensor can be used for several functions, including mapping the environment based on information of distance of all object on the workspace, localizing the mobile robot on the global map based on perception of the environment and avoiding collision during the operation by detecting an obstacle or object along the robot way.

Some popular distance sensors used in mobile robot application are including Kinect, Hokuyo UTM-30LX, and RPLidar. Some research studies and applications have been published related to the implementations of these sensors. The Kinect sensor was used by Peter *et al.* for 3D mapping on the indoor application [1]. Meanwhile, this sensor was also used by Jagdish *et al.* for hand tracking study and recognizing the center of the hand [2]. Moreover, Midriem *et al.* used Kinect sensor for detecting and calculating the distance of specific object for weapon system application [3]. On the other hand, Nicolas *et al.* have used Hokuyo UTM-30LX sensor, for detecting an obstacle on electrical wire routes [4], while Ji *et al.* using it for a real-time method for depth enhanced visual odometry [5]. Meanwhile, RPLidar sensor has been used by Marni *et al*. for scanning and mapping on the indoor environment [6]. Similarly, Mirna *et al.* were also used this sensor on the autonomous mobile robot for mapping the environment [7].

---
* Corresponding Author. Tel: +62 22 250 3055
  E-mail address: midr001@lipi.go.id

In this paper, Kinect, Hokuyo UTM-30LX, and RPLidar sensor will be analyzed and discussed to see the actual performance of these three sensors to detect two different types of objects, non-transparent and transparent objects in the various tested distance. A software application created by C programming has been developed to utilize data from each sensor. The experiment results in this paper can be used as references to select the right distance sensor for further applications.

## II. Research Method

### A. Distance Sensors

Kinect, Hokuyo UTM-30LX, and RPLidar sensors discussed in this paper are presented in Figure 1, Figure 2, and Figure 3. The Kinect sensor in Figure 1 is a sensor used for Xbox 360 console. This sensor consists of four main components, RGB camera, 3d depth sensor, microphone array and motorized tilt made by Microsoft [8]. The depth sensor component on the Kinect can be used to localize objects on three-dimensional coordinate frames, i.e., X, Y, and Z on meter unit. Some related specifications of this Kinect sensor can be seen in Table 1. Based on the specification list in Table 1, this sensor has effective distance measurement from 0.8 to 4.0 meter.

Meanwhile, Hokuyo UTM-30LX is one of Light Detection and Ranging (LiDAR) technologies that can measure object distance and bearing by emitting laser signal into the measured object. After that, the reflected laser signal will be read for calculating the object distance. The object distance is calculated based on Time of Flight (ToF) of the laser signal. Some related specifications of this Hokuyo UTM-30LX sensor are listed in Table 2 [9]. According to this list, the effective distance measurement capable by this sensor is between 0.1 and 30 meters, with accuracy about ± 30mm.

On the other hand, RPLidar sensor was designed as a low-cost two-dimensional laser scanner compare to the existed commercial laser scanner. These sensor measures object distance using triangulation principle such as illustrated in Figure 4. Generally speaking, RPLidar has three main components, signal transmitter system, vision acquisition system, and a motor system that spins these two previous components. The transmitter component emits the modulated laser infrared signal that will hit an object. After that, the vision acquisition system will catch the reflected infrared signal from the object, and the distance will be calculated based on the triangulation principle. The general specification of this RPLidar sensor can be seen in Table 3 [10]. Based on the data, it shows that this sensor has angular span 360 degrees with less than one-degree resolution. The effective measuring distance is about 0.2 to 6 meter.

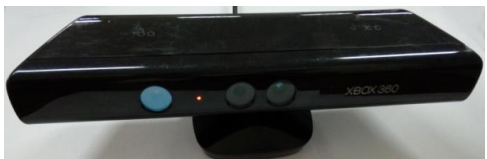

Figure 1. Kinect

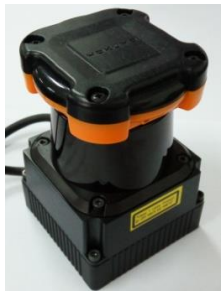

Figure 2. Hokuyo UTM-30LX

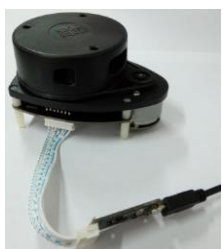

Figure 3. RPLidar

Table 1.
Kinect specification

| Parameter | Specification |
| --- | --- |
| Effective measurement distance | 0.8 - 4.0 meter |
| Measurement range angle | 43° on vertical |
| | 57° on horizontal |
| Accuracy | N/A |
| Tilt | ±27° |
| Frame rate | 30 frames per second (FPS) |

Table 2.
Hokuyo UTM-30LX specification

| Parameter | Specification |
| --- | --- |
| Effective measurement distance | 0.1 – 30 meter |
| Accuracy | 0.1 – 10m : ± 30mm |
| Scan speed | 25 msec/scan |
| Scan angle | 270° |
| Angular resolution | 0.25° |

Table 3.
RPLidar specification

| Parameter | Specification |
| --- | --- |
| Distance range | 0.2 - 6 meter (typically) |
| Distance resolution | < 0.5 mm or < 1% of the distance |
| Angular range | 0 - 360° |
| Angular resolution | ≤ 1° |
| Scan Rate | Min: 1 Hz, Max: 10 Hz |

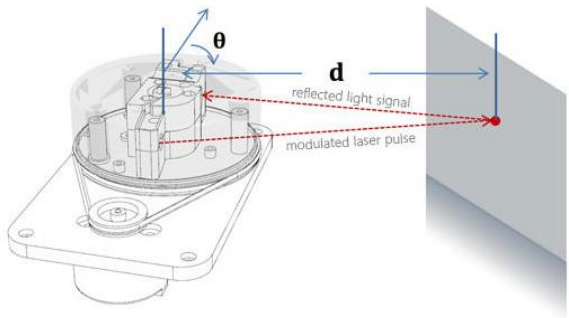

Figure 4. Distance measurement illustration of the RPLidar sensor with the triangulation principle

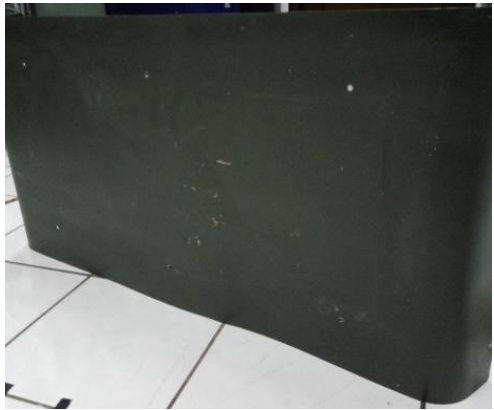

Figure 5. Metal plate object

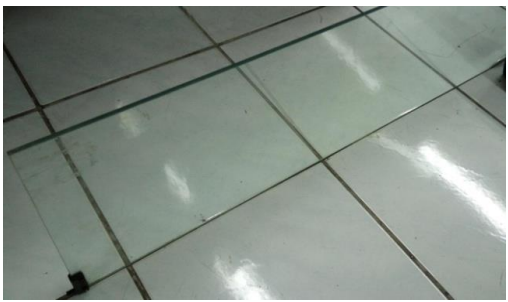

Figure 6. Transparent glass object

## B. Experimental methods and measuring techniques

In this study, experimental testing has been conducted on these three different sensors, i.e., Kinect, Hokuyo UTM-30LX, and RPLidar. This experiment is performed to review the actual performance of these sensors on measuring object distance. The objects used in this experiment consist of two different types, non-transparent object, and transparent object. A dark green metal plate object with a thickness of 0.8 mm shown in Figure 5 is used to represent the non-transparent object. Meanwhile, the 5 mm thick glass used to represent the transparent object can be seen in Figure 6.

The main purpose of the non-transparent object experiment is to observe the performance of these sensors to measure the object distance in the various tested distance on the same object. More specifically, this experiment will observe the measurement variance of each sensor on the same object and same distance, the effect of the measurement distance into measurement error of each sensor and the actual range measurement of each sensor. In this experiment, the same object is measured on the distance 100 to 3000 mm, with every 100 mm iteration. The process layout of this experiment can be illustrated as in Figure 7.

On the other hand, the transparent object experiment is conducted to observe the sensitivity of each sensor to detect a transparent object. Moreover, this experiment is also observing the effect of the transparent object to the distance measurement result of the non-transparent object behind it. In other words, this experiment will observe the refraction effect of the glass to the distance measurement result of each sensor. This experiment is performed by placing the glass in front of the sensors in various distances, 100, 200, 300, 400, and 500 mm and also placing the dark metal plate behind the glass on 2000 mm in front of the sensors. The process layout of this second experiment can be illustrated as in Figure 8.

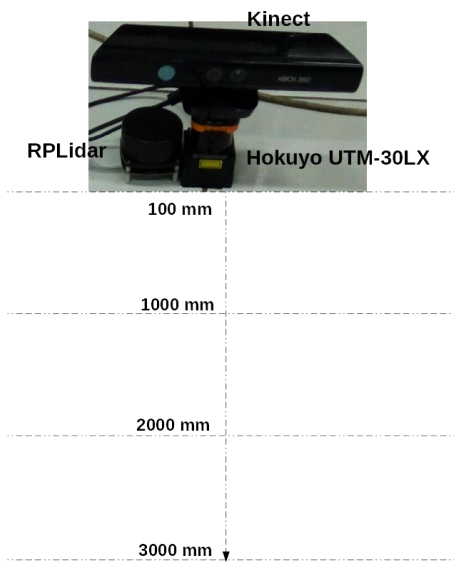

Figure 7. Design plans for the distance sensors to calculate the metal plate object

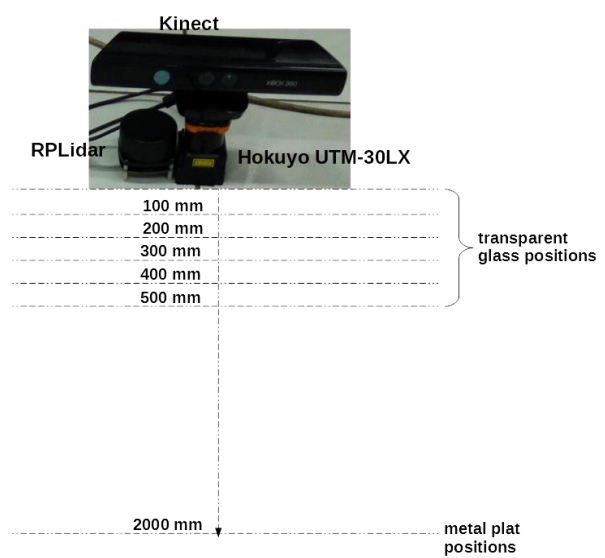

Figure 8. Design plans for the distance sensors to calculate the transparent glass object

## C. Obtaining data from the sensors

Figure 9 shows the flowchart of the software application for obtaining distance measurement data from the sensors. This application is developed on C programming platform using Visual Studio IDE which can be seen in Figure 10.

The program records all obtained distance data and processing time of each sensor on the text file for each experiment. All sensors are connected to the PC through USB connection. The data that is accessed by the program on this experiment represents the distance data on of the object in the direction perpendicular to the center of each sensor. The corresponding data for Kinect sensors is the data on the pixel 320x240 on the depth image frame. The corresponding data for the Hokuyo UTM-30LX is the distance data on the 540th step. Meanwhile, on the RPLidar sensor, the accessed data is the data on the scanning angle zero degrees.

## III. Result and Discussion

The results of these experiments were evaluated to determine the actual performance of each sensor based on these actual result.

### A. Experiment result in the non-transparent object

The measurement results of these three sensors on the non-transparent object placed on 100 up to 3000 mm on distance can be seen in Table 4. According to these results, it can be summarized in Table 5, the actual distance measurement range that can be covered by these sensors on the experiment. Comparing this result with the specification in Table 2, it was shown that in reality the actual measurement of the measuring range of the sensor, particularly the Kinect sensor, performed under its manufacturing specifications. It might be affected by various things, including the lighting effects in the room when the testing process performed, the type of the object being detected and also the color of the object being detected.

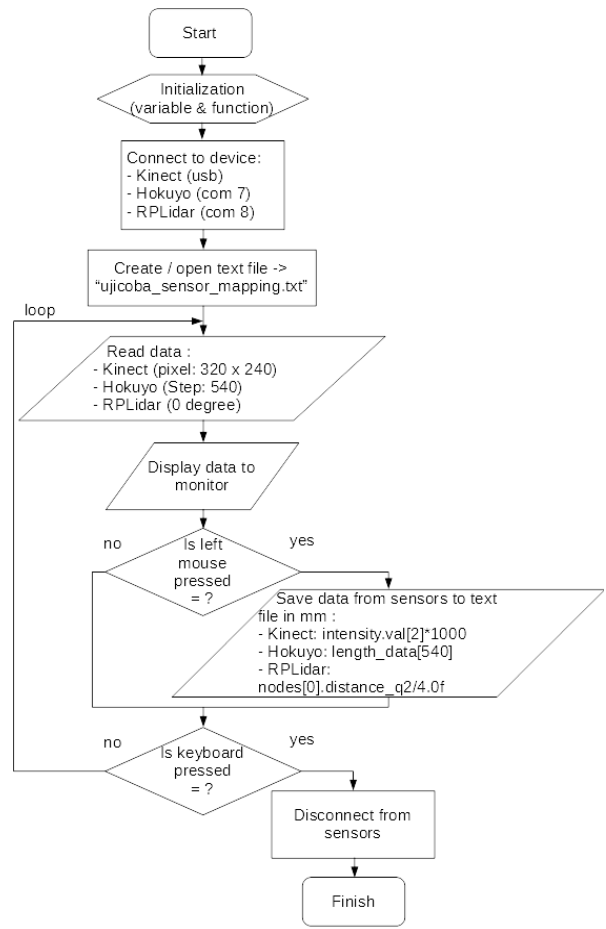

Figure 9. Flowchart of the software application for obtaining distance measurement from the sensors

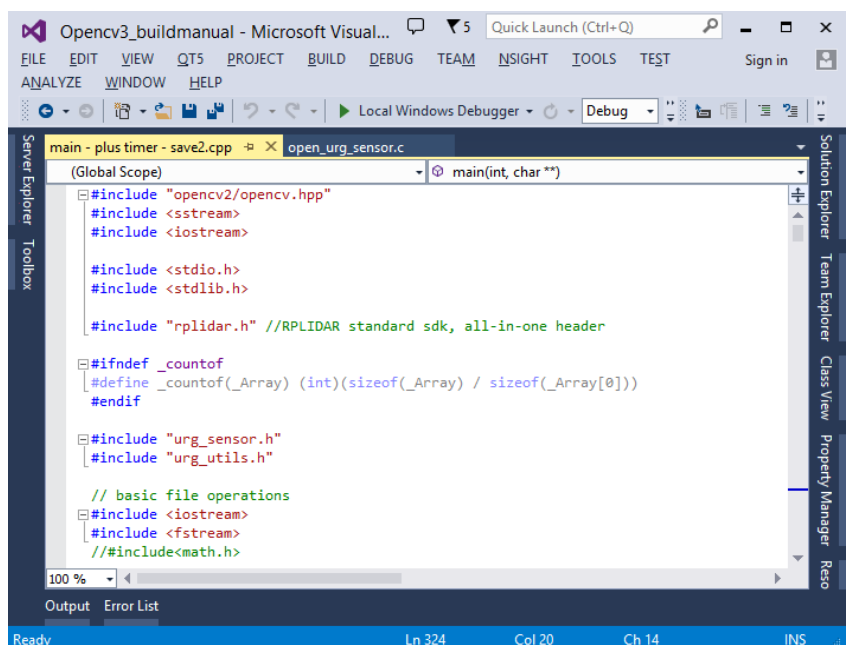

Figure 10. Visual Studio IDE

Table 4.
The measurement results of the three sensors on the non-transparent object

| Actual distance | Distance measurement average | | |
|---|---|---|---|
| | **Kinect** | **Hokuyo** | **RPLidar** |
| 100 | N/A | 127.6 | 131.575 |
| 200 | N/A | 223.2 | 234.275 |
| 300 | N/A | 321.4 | 343.025 |
| 400 | N/A | 419.7 | 443 |
| 500 | 509.8 | 521.6 | 543.2 |
| 600 | 606 | 620.1 | 647.85 |
| 700 | 706.6 | 718.3 | 752.65 |
| 800 | 803.6 | 815.9 | 865 |
| 900 | 901 | 914.9 | 975.125 |
| 1000 | 1003 | 1017.1 | 1079.9 |
| 1100 | 1098.8 | 1113.5 | 1183.075 |
| 1200 | 1202 | 1212.2 | 1294.8 |
| 1300 | 1294 | 1312.4 | 1399.85 |
| 1400 | 1403.5 | 1422.8 | 1505.8 |
| 1500 | 1499.7 | 1524 | 1606.525 |
| 1600 | 1590.4 | 1623.5 | 1730.925 |
| 1700 | 1697.2 | 1722 | 1840.275 |
| 1800 | 1786.9 | 1823.7 | 1944.275 |
| 1900 | N/A | 1926.5 | 2055.05 |
| 2000 | N/A | 2020.4 | 2149.525 |
| 2100 | N/A | 2123.6 | 2280.2 |
| 2200 | N/A | 2223.2 | 2380.425 |
| 2300 | N/A | 2323.7 | 2493.025 |
| 2400 | N/A | 2420.2 | 2603.325 |
| 2500 | N/A | 2526.5 | 2726.875 |
| 2600 | N/A | 2627 | 2810.65 |
| 2700 | N/A | 2728.4 | 2969.7 |
| 2800 | N/A | 2827.7 | 3090.9 |
| 2900 | N/A | 2930.6 | 3188.125 |
| 3000 | N/A | 3025.6 | 3317.5 |

Table 5.
The actual distance measurement range that can be covered by the three sensors on the experiment

| Sensor | min | max |
|---|---|---|
| Kinect | > 400 mm | < 1900 mm |
| Hokuyo | 0 ≤ 100 mm | ≥ 3000 mm |
| RPLidar | 0 ≤ 100 mm | ≥ 3000 mm |

Table 6.
The correlation coefficient between the measurement error and the distance of the object from the sensors

| Sensor | Correlation coefficient |
|---|---|
| Kinect | -0.808 |
| Hokuyo | 0.435 |
| RPLidar | 0.97 |

Table 7.
The model equations for the three sensors to distance measurement

| Linear Regression Equation Model | R-Sq Value |
|---|---|
| E_Kinect = 14.78 - 0.013 tested_distance | 64.2 % |
| E_RPLidar = - 7.89 + 0.093 tested_distance | 94.1 % |

Table 8.
Descriptive statistical summary of measurement error resulted by Hokuyo sensor

| Statistical parameter | Hokuyo |
|---|---|
| Observation Number (N) | 300 |
| Mean | 21.94 |
| StDev | 5.67 |
| Variance | 32.11 |
| Minimum value | 4 |
| Maximum value | 37 |
| Range | 33 |

The errors of the distance measurements from each sensor on this experiment were presented in Figure 11a, Figure 11b, and Figure 11c. According to the graph on the Figure 11b, it was shown that the measurement error of the Hokuyo UTM-30LX on all tested distance did not indicate any particular significant trend. In contrast, the graphs in Figure 11a and 11c showed particular trends of the measurement error of the Kinect and RPLidar sensors. These graphs indicated negative and positive slope, respectively.

The correlation between tested actual distance and measurement error results on each sensor could also be evaluated using the Pearson Correlation method [11]. The correlation coefficient between the measurement error and the distance of the object from the sensors could be seen in Table 6.

Based on correlation coefficient shown in Table 6, it confirmed the rapid conclusion based in Figure 11 that the error on Hokuyo sensor had no strong correlation with tested distance, while Kinect and RPLidar sensors had negative and positive correlations, respectively.

More specifically, since Kinect and RPLidar sensors had strong correlations between measurement error and tested distance, these correlations can be modeled using fitted linear regression model. The linear fitted line plot regression of these correlations on the Kinect and RPLidar sensors could be seen in Figure 12, and the model equations for these sensors were listed in Table 7.

According to Figure 12, it was shown that linear regression model on the RPLidar sensor could result in a well-fitting model, while on Kinect sensors, the linear regression model results in a less-fitting model. The R-squared values also indicated that the estimated model for correlation error on the RPLidar sensor was much closer to perfect model with R-squared value 94.1% compared to the model for correlation error on the Kinect sensor with R-squared value 64.2%.

Meanwhile, since there was no significant correlation between tested distance and measurement error resulted by Hokuyo sensor, the measurement error can be evaluated and modeled as a normally distributed error model. Table 8 showed the statistical summary of the measurement error resulted by Hokuyo sensor on this experiment.

The normality of the error of this sensor was evaluated based on the residual distribution of this error. Figure 13 demonstrated the residual plots of the measurement error on the Hokuyo sensor. According to the normal probability plot in the figure, it indicated that the residual error was well-fitted being modeled as a normal distribution model. Moreover, versus fits the plot, and versus order plot did not show any particular trend. It indicated that the error was the independent one to the others in any observation. Also, the normality of this residual error could be seen visually on the histogram on the figure. Thus, based on this observation and evaluations, the measurement error resulted by Hokuyo sensor was well-fitted to be modeled as normally distributed error with mean 21.94 and variance 32.11. In this experiment, apart from the evaluation of the error of the distance measurement results, the processing time required by each sensor to process one cycle measurement was also evaluated.

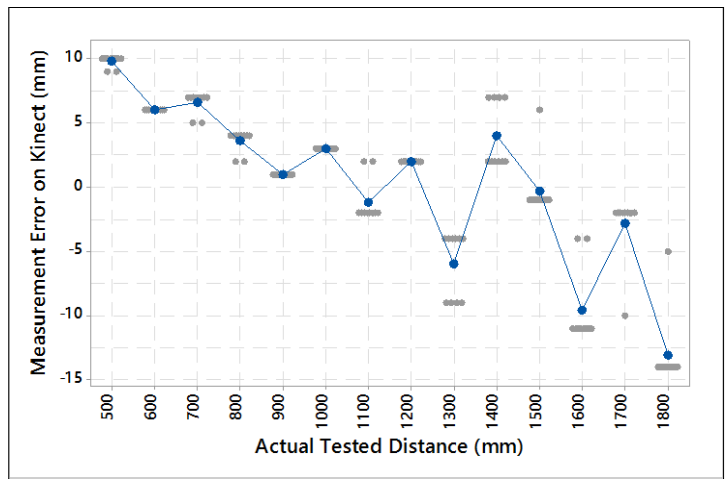

(a)

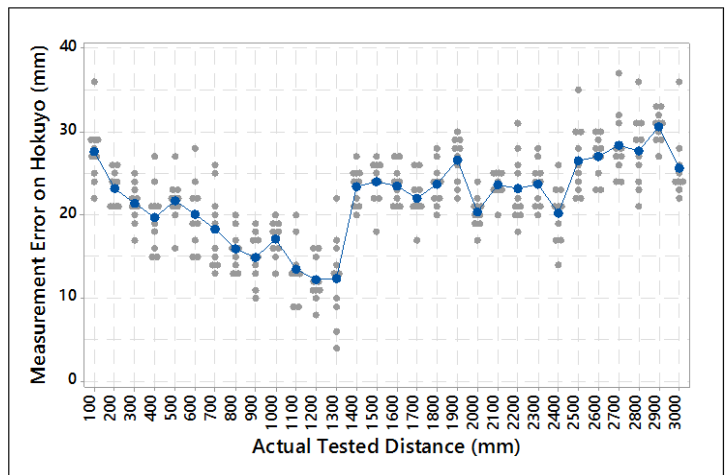

(b)

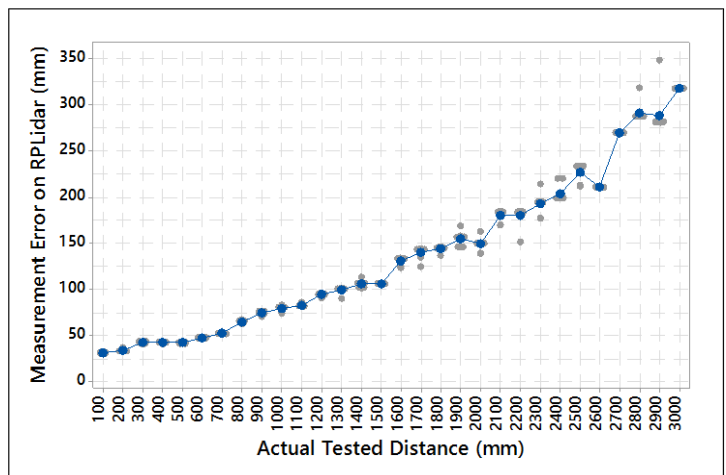

(c)

Figure 11. The errors of the distance measurements from each sensor on this experiment; (a) Kinect; (b) Hokuyo; and (c) RPLidar

On Kinect sensor, every one cycle, this sensor measure object distance covering the three-dimensional frame with scope 43° in the vertical direction and 57° in the horizontal direction. It was represented by 320 x 480 pixels of depth image frame. Conversely, Hokuyo UTM-30LX only covers two-dimensional plane distance measurement. One cycle on this sensor covers the measurement of the 270 degrees scope.

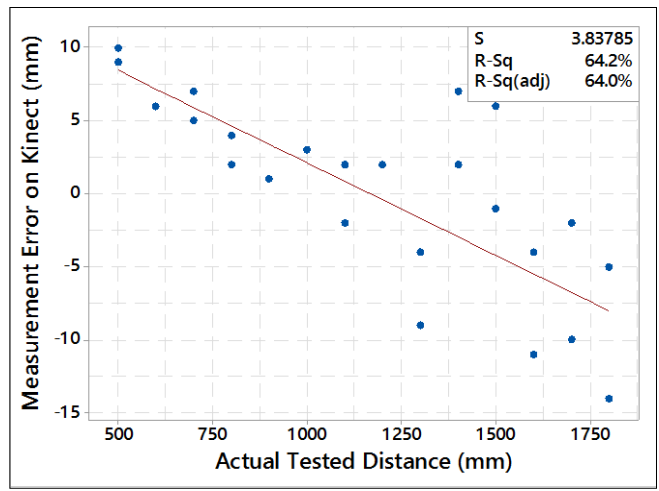

(a)

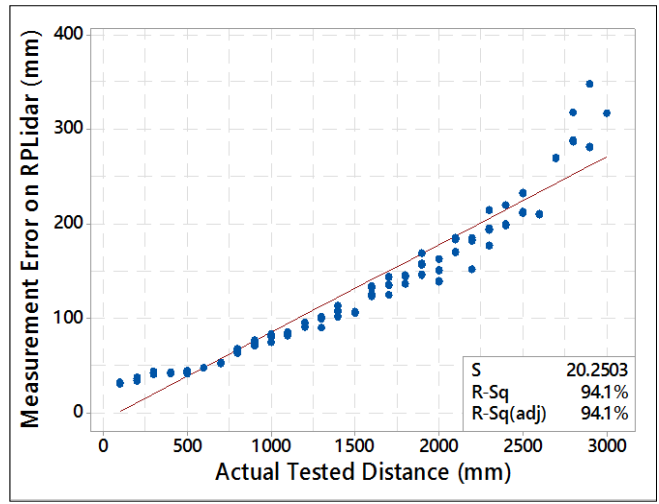

(b)

Figure 12. The linear fitted line plot regression of these correlations on the Kinect and RPLidar sensors; (a) Kinect and (b) RPLidar

Similar to the Hokuyo UTM-30LX, RPLidar was also cover two-dimensional plane distance measurement. This sensor could cover the scope of 270 degree measurement of distance on everyone measurement cycle. Figure 14 showed the processing times required by these sensors on measuring object on the various distances on every one cycle measurement.

Based on Figure 14, it could be rapidly seen that no significant trend showed the correlation between processing time and tested distance on each sensor measurement results. More specifically, the correlation between processing time and tested distance on each sensor measurement results could be evaluated using Pearson Correlation method. The correlation coefficient between the processing time and distance measurements from three tested sensors could be seen in Table 9. From the Table 9, it could be seen that the correlation coefficient between measurement time and measurement distance was relatively small for each sensor, so generally speaking, the tested distances on the measurement processes had no significant effect on the measurement processing time.

The average measurement processing time of one cycle of the three sensors could be seen in Table 10. Practically, one cycle, RPLidar sensor needs significantly much longer processing time than the two other sensors. Meanwhile, Hokuyo UTM-30LX sensor required the fastest processing time compared to the others.

Table 9.
The correlation coefficient between the processing time and distance measurements

| Sensor  | Correlation coefficient |
|---------|-------------------------|
| Kinect  | -0.056                  |
| Hokuyo  | -0.112                  |
| RPLidar | -0.093                  |

Table 10.
The average measurement processing time

| Sensor  | The average measurement processing time (ms) |
|---------|----------------------------------------------|
| Kinect  | 13.49165                                     |
| Hokuyo  | 3.843492                                     |
| RPLidar | 125.7597                                     |

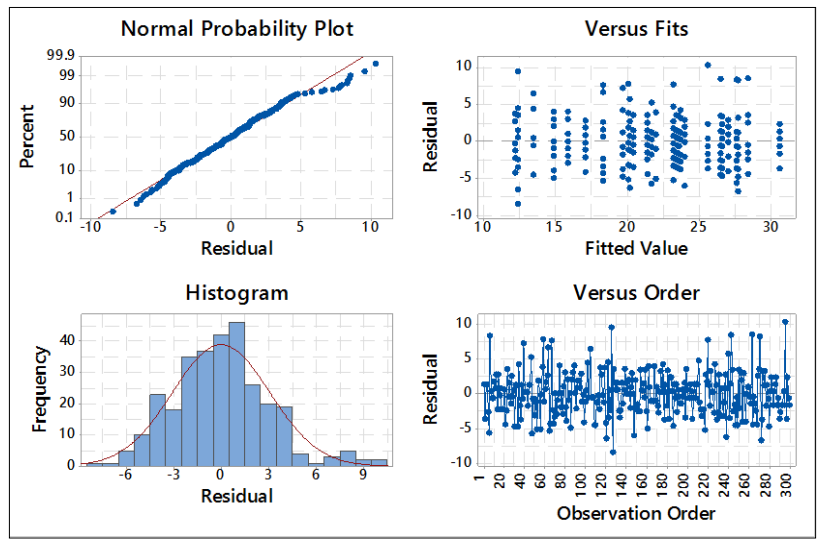

Figure 13. The residual plots of the measurement error on the Hokuyo sensor

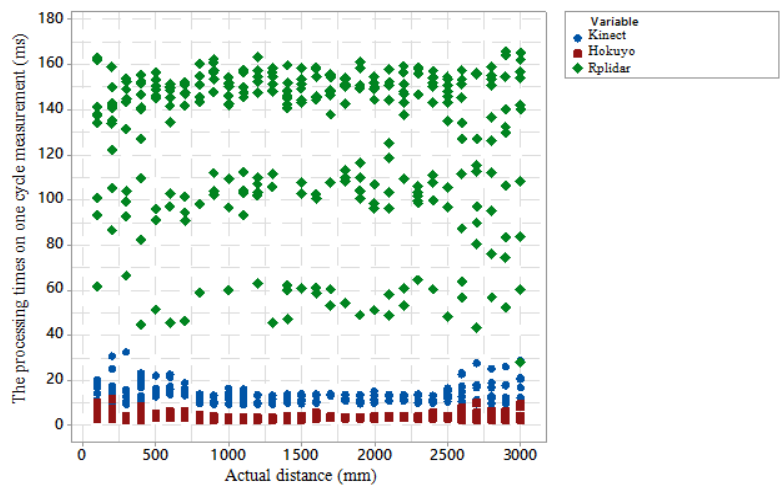

Figure 14. The processing times required by these sensors on every one cycle measurement

### B. Experiment result on the transparent object

Table 11 presented the measurement results of the transparent object's distance using Kinect, Hokuyo UTM-30LX, and RPLidar sensors. It could be seen clearly that the Kinect sensor can adequately detect the transparent glass at its effective distance (i.e., $\geq$ 500 mm), as the Hokuyo UTM-30LX sensor could only detect the transparent glass for measuring the distance of more than equal to 200 mm. By contrast, the RPLidar sensor cannot detect the transparent glass at all, for any tested distance. When the transparent glass was not detected, the detected object is the object behind the glass, which was the dark green metal plate. This plate was placed at a distance of 2000 mm from the sensor. Table 12 showed the comparison of distance measurement result of the metal plate placed on 2000 mm, with glass existing and without glass existing for Hokuyo UTM-30LX and RPLidar sensors. The comparison of the error of these measurement results could be seen in Figure 15a and 15b. Based on Table 12 and Figure 15, it showed that the average error of measurement object by the sensor Hokuyo through transparent glass was -54 mm, and without a transparent glass was 20 mm.

Table 11.
The measurement results of the transparent object's distance

| Transparent glass position (mm) | Metal plate position (mm) | The measurement results (mm) | | |
|---|---|---|---|---|
| | | Kinect | Hokuyo | RPLidar |
| 100 | | N/A | 1946.4 | 2087.075 |
| 200 | | N/A | 232.4 | 2101.05 |
| 300 | 2000 | N/A | 340.3 | 2138.05 |
| 400 | | N/A | 420.1 | 2146.125 |
| 500 | | 508.5 | 559.4 | 2150.25 |

Table 12.
The comparison of distance measurement result of the metal plate, with and without glass existing

| Metal plate distance (mm) | Transparent glass distance (mm) | The measurement results (mm) | | | |
|---|---|---|---|---|---|
| | | With transparent glass intercession | | Without transparent glass intercession | |
| | | **Hokuyo** | **RPLidar** | **Hokuyo** | **RPLidar** |
| 2000 | 100 | 1945 | 2095.25 | 2021 | 2151 |
| | | 1947 | 2094.5 | 2020 | 2139.5 |
| | | 1943 | 2094.5 | 2019 | 2150.5 |
| | | 1946 | 2082 | 2019 | 2150.25 |
| | | 1949 | 2081.75 | 2021 | 2150.5 |
| | | 1945 | 2069.25 | 2020 | 2150.5 |
| | | 1945 | 2082.25 | 2022 | 2162.75 |
| | | 1946 | 2082 | 2017 | 2150.75 |
| | | 1952 | 2094.5 | 2024 | 2150.5 |
| | | 1946 | 2094.75 | 2021 | 2139 |

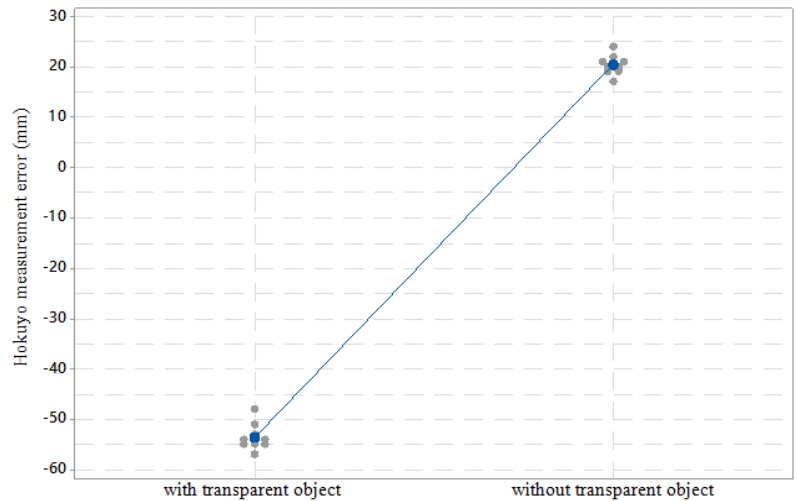

(a)

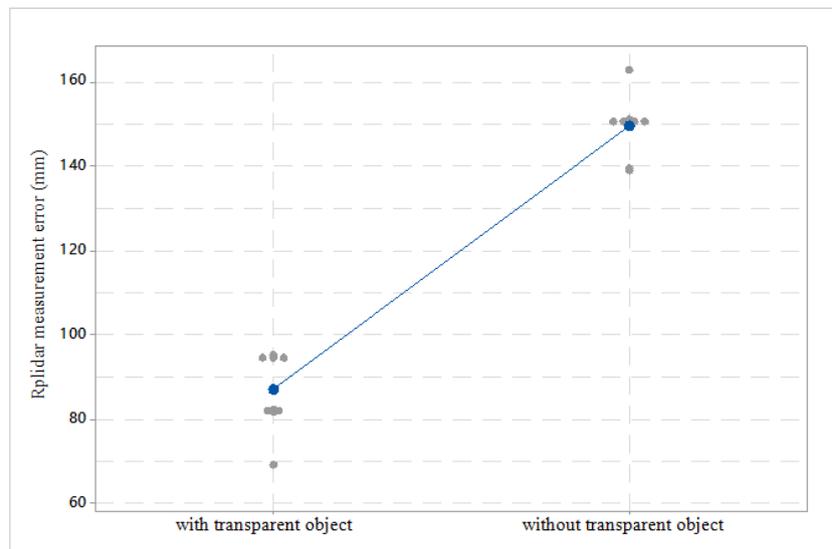

(b)

Figure 15. Error measurement result of the Hokuyo and RPLidar sensors with or without glass existing (a). Hokuyo; (b). RPLidar

Meanwhile, the average error of measurement object by the sensor RPLidar through transparent glass was 88 mm, and without a transparent glass was 149 mm. These results showed roughly that the glass appearance affects the measurement result of these sensors. This effect was as a result of diffraction phenomenon. Statistically, this comparison was also can be tested using the 2-sample t-test method. The result of t-test could be seen in Table 13. The comparison of the Hokuyo sensor results T-value as -74.5, and the RPLidar sensor results T-value as -17.81. The comparison results on both two sensors showed the high T-Values. It confirmed that even though the glass cannot detect by the sensors, the glass appearance significantly affects the distance measurement result of the Hokuyo and RPLidar sensors.

Table 13.
The comparison of distance measurement result of the metal plate, with and without glass existing

| Sensor | Condition | N | Error Average | StDev | Average different | T-Value |
|---|---|---|---|---|---|---|
| Hokuyo | With transparent object | 10 | -53.6 | 2.5 | -74 | -74.5 |
| | Without transparent object | 10 | 20.4 | 1.9 | | |
| RPLidar | With transparent object | 10 | 211.98 | 4.35 | -62.45 | -17.81 |
| | Without transparent object | 10 | 274.43 | 2.08 | | |

## IV. Conclusion

An experimental testing on the Kinect, Hokuyo UTM-30LX, and RPLidar sensors had been conducted to test the actual performance of these three sensors using two different types of objects, non-transparent object, which was a dark green metal plate, and a transparent object, which was a 5 mm thick transparent glass. The Kinect sensor could detect objects with a minimum distance of > 400 mm, while the Hokuyo and RPLidar sensors already could detect an object in the distance about 100 mm (i.e., The minimum distance tested in this experiment). While the Hokuyo UTM-30LX and RPLidar sensors could detect the object on the distance up to 3000 mm (i.e., The minimum distance tested in this experiment), on this experiment, the Kinect sensor could only detect the object on maximum distance 1900 mm. Considering the various distance measurement in this experiment, the results showed that the Hokuyo UTM-30LX did not have a strong correlation between the measurement errors and the measurement distance tested. More specifically, the normality indicated that the error resulted by this sensor is well-fitted modeled as a normally distributed error with mean 21.94 mm and variance 32.11. In contrast, the measurement errors resulted by Kinect and RPLidar sensors had strong correlations with the measurement distance tested the error on the Kinect sensor had a strong negative correlation, while the error resulted by RPLidar sensor had a strong positive correlation with the tested distance. The performance of these three sensors for detecting a transparent object tested in this experiment (i.e., 5 mm thick transparent glass), showed that the Kinect sensor could detect the transparent object on its effective range measurement, and Hokuyo UTM-30LX could detect the transparent object in the distance more than equal to 200 mm. On the other hand, the RPLidar sensor cannot detect the transparent object at all tested distance. While the transparent object was not detected by the sensors, this object still significantly affected the measurement result of the sensor when measuring the distance of the object behind this transparent object. Lastly, the performance of these three sensors regarding processing time, it was shown that the Hokuyo UTM-30LX had the fastest processing time significantly, and the RPLidar had the slowest processing time significantly, while the processing time of Kinect sensor was in between both. These processing times were not significantly affected by various tested distance measurement.

## Acknowledgement

Authors would like to thank to the Research Centre for Electrical Power and Mechatronics - Indonesian Institute of Sciences (LIPI) that has supported this research and all those who have helped in conducting this research.

## References

[1] P. Henry *et al.*, "RGB-D mapping: Using Kinect-style depth cameras for dense 3D modelling of indoor environments," *The International Journal of Robotics Research,* vol. 31, no. 5, pp. 647-663, 2012.
[2] J. L. Raheja *et al.*, "Tracking of Fingertips and Centers of Palm Using KINECT," in *Third International Conference on Computational Intelligence, Modelling & Simulation*, Langkawi, 2011.
[3] M. Mirdanies *et al.*, "Object Recognition System in Remote Controlled Weapon Station using SIFT and SURF Methods," *Journal of Mechatronics, Electrical Power, and Vehicular Technology*, vol. 4, no. 2, pp. 99-108, 2013.
[4] N. Pouliot *et al.*, "LineScout power line robot: Characterization of a UTM-30LX LIDAR system for obstacle detection," in *IEEE/RSJ International Conference on Intelligent Robots and Systems*, Vilamoura, 2012.
[5] J. Zhang *et al.*, "A real-time method for depth enhanced visual odometry," *Auton. Robots,* vol. 41, no. 1, pp. 31–43, Jan. 2017.
[6] A. M. Markom *et al.*, "Indoor Scanning and Mapping using Mobile Robot and RP Lidar*," Int'l Journal of Advances in Mechanical & Automobile Engg. (IJAMAE)*, vol. 3, no. 1, pp. 42-47, 2016.
[7] A. M. Markom *et al.*, "A mapping mobile robot using RP Lidar Scanner," in *IEEE International Symposium on Robotics and Intelligent Sensors (IRIS),* Langkawi, 2015.
[8] Microsoft, "Kinect for Windows Sensor Components and Specifications," Microsoft, US, 2016.
[9] Hokuyo, "Scanning Laser Range Finder UTM-30LX/LN Specification," Hokuyo Automatic CO., LTD, 2012.
[10] RoboPeak Team, "RPLIDAR Low Cost 360 degree 2D Laser Scanner (LIDAR) System," 2014.
[11] J. Adler dan I. Parmryd, "Quantifying colocalization by correlation: The Pearson correlation coefficient is superior to the Mander's overlap coefficient," *Cytometry Part A*, vol. 77A, no. 8, pp. 733-742, 2010.